\documentclass[final]{cvpr}
\usepackage{times}
\usepackage{epsfig}
\usepackage{graphicx}
\usepackage{amsmath}
\usepackage{amssymb}

\usepackage{amsmath}
\usepackage{array}

\usepackage{graphicx}
\usepackage{algorithm}
\usepackage{algorithmic}
\usepackage{booktabs}
\usepackage{multirow}
\usepackage{amsfonts}
\usepackage{amssymb}

\usepackage{nicefrac}
\usepackage{url}

\newcommand{\R}{\mathbb{R}}

\makeatletter
\newcommand{\ssymbol}[1]{$^{\@fnsymbol{#1}}$}
\makeatother

\usepackage[pagebackref=true,breaklinks=true,colorlinks,bookmarks=false]{hyperref}

\def\confYear{CVPR 2021}

\begin{document}

\title{Exploring and Distilling Posterior and Prior Knowledge for \\ Radiology Report Generation}

\author{Fenglin Liu\textsuperscript{1}, Xian Wu\textsuperscript{2}, Shen Ge\textsuperscript{2}, Wei Fan\textsuperscript{2}, Yuexian Zou\textsuperscript{1,3}\\
\textsuperscript{1}ADSPLAB, School of ECE, Peking University\\
\textsuperscript{2}Tencent Medical AI Lab  \ \ \textsuperscript{3}Peng Cheng Laboratory \\
{\tt\small \{fenglinliu98, zouyx\}@pku.edu.cn, {\tt \{kevinxwu, shenge, Davidwfan\}@tencent.com}}
}

\maketitle

\begin{abstract}

Automatically generating radiology reports can improve current clinical practice in diagnostic radiology. On one hand, it can relieve radiologists from the heavy burden of report writing; On the other hand, it can remind radiologists of abnormalities and avoid the misdiagnosis and missed diagnosis. Yet, this task remains a challenging job for data-driven neural networks, due to the serious visual and textual data biases. To this end, we propose a Posterior-and-Prior Knowledge Exploring-and-Distilling approach  (PPKED) to imitate the working patterns of radiologists, who will first examine the abnormal regions and assign the disease topic tags to the abnormal regions, and then rely on the years of prior medical knowledge and prior working experience accumulations to write reports. Thus, the PPKED includes three modules: Posterior Knowledge Explorer (PoKE), Prior Knowledge Explorer (PrKE) and Multi-domain Knowledge Distiller (MKD). In detail, PoKE explores the posterior knowledge, which provides explicit abnormal visual regions to alleviate visual data bias; PrKE explores the prior knowledge from the prior medical knowledge graph (medical knowledge) and prior radiology reports (working experience) to alleviate textual data bias. The explored knowledge is distilled by the MKD to generate the final reports. Evaluated on MIMIC-CXR and IU-Xray datasets, our method is able to outperform previous state-of-the-art models on these two datasets.

\end{abstract}

\begin{figure}[t]

\centering
\begin{center}
\includegraphics[width=1\linewidth]{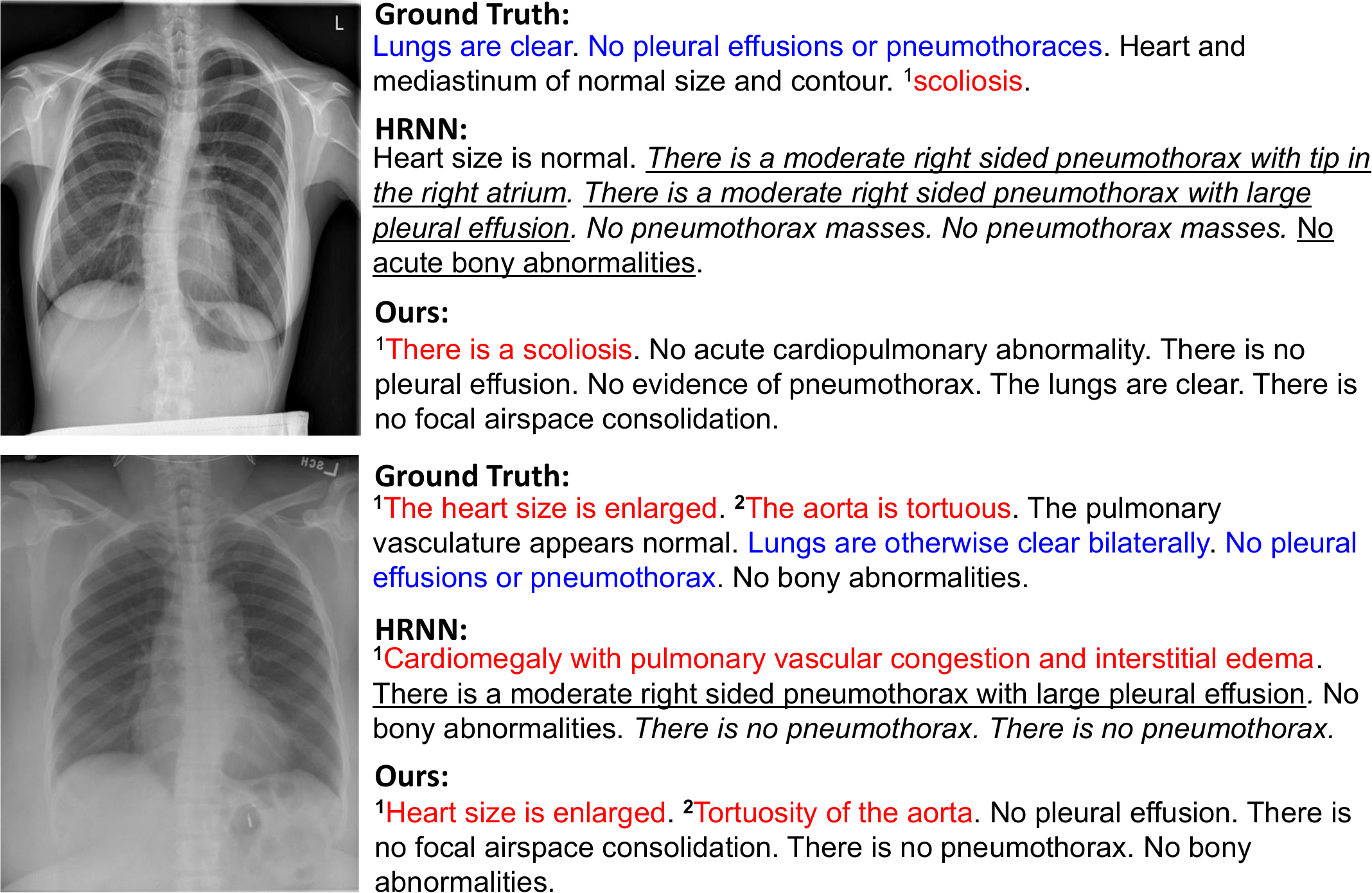}
\end{center}
\caption{Two examples of ground truth reports and reports generated by HRNN \cite{Krause2017Hierarchical} and our method. The Red colored text indicates the abnormalities in reports. The Blue colored text stands for the similar sentences used to describe the normalities in ground truth reports. There are notable data bias and the HRNN fails to depict some rare but important abnormalities and generates some error sentences (Underlined text) and repeated sentences (Italic text).}
\label{fig:introduction}
\end{figure}

\section{Introduction}
Medical images like radiology and pathology images are widely-used in disease diagnosis and treatment \cite{Delrue2011Difficulties}. Given a radiology image, radiologists first examine both the normal and abnormal regions and then use the learned medical knowledge and accumulated working experience to write a coherent report to note down the findings \cite{WHO2004Neurology,goergen2013evidence,Li2019Knowledge}. Given the large volume of radiology images, writing reports become a heavy burden for radiologists. Furthermore, for less experienced radiologists, some abnormalities in radiology images may be ignored and consequently not included in the reports \cite{Brady2012DiscrepancyAE}. To relieve radiologists from such heavy workload and remind inexperienced radiologists of abnormalities, automatically generating radiology reports becomes a critical task in clinical practice.

In recent years, automatic radiology report generation has attracted extensive research interests \cite{Zhang2020When,Li2018Hybrid,Wang2018TieNet,Jing2019Show,Chen2020Generating,liu2021Contrastive}. Most existing methods, like \cite{Jing2018Automatic,Xue2018Multimodal,Yuan2019Enrichment} follow the standard image captioning approaches and employ the encoder-decoder framework, e.g., CNN-HRNN  \cite{Jing2018Automatic,Liang2017Hierarchical}. In the encoding stage, the image features are extracted by CNN from the entire image; In the decoding stage, the whole report is generated by HRNN. However, directly applying image captioning approaches to radiology images   has the following problems: 1) Visual data deviation: the appearance of normal images dominate the dataset over that of abnormal images \cite{Shin2016Learning}. 
As a result, this unbalanced visual distribution would distract the model from accurately capturing the features of rare and diverse abnormal regions. 2) Textual data deviation: as shown in Figure~\ref{fig:introduction}, in a report, radiologists tend to describe all the items in an image, making the descriptions of normal regions dominate the entire report. Besides, many similar sentences are used to describe the same normal regions. With this unbalanced textual distribution, training with such dataset makes the generation of normal sentences dominant \cite{Jing2019Show,Xue2018Multimodal,Yuan2019Enrichment,liu2021Contrastive}, disabling the model to describe specific crucial abnormalities. In brief, as shown in Figure~\ref{fig:introduction}, the widely-used HRNN \cite{Krause2017Hierarchical} generates some repeated sentences of normalities and fails to depict some rare but important abnormalities.

To ensure these rare but important abnormal regions captured and described, the urgent problem is to alleviate such serious data deviation problem \cite{Shin2016Learning,Li2018Hybrid,Jing2019Show,Li2019Knowledge,Zhang2020When,liu2021Contrastive}. In our work, we propose the Posterior-and-Prior Knowledge Exploring-and-Distilling (PPKED) framework, which imitates the radiologists' working patterns to alleviate above problems. Given a radiology image, radiologists will \textit{examine the abnormal regions} and  \textit{assign the disease topic tags to the abnormal regions}; then \textit{accurately write a corresponding report} based on years of \textit{prior medical knowledge} and \textit{prior working experience} accumulations \cite{goergen2013evidence,Delrue2011Difficulties,Li2019Knowledge}. In order to model above working patterns, the PPKED introduces three modules, i.e., Posterior Knowledge Explorer (PoKE), Prior Knowledge Explorer (PrKE) and Multi-domain Knowledge Distiller (MKD). The PoKE could alleviate visual data deviation by \textit{extracting the abnormal regions} based on the input image; The PrKE could alleviate textual data deviation by encoding the prior knowledge, including the prior radiology reports (i.e., \textit{prior working experience}) pre-retrieved from the training corpus and the prior medical knowledge graph (i.e., \textit{prior medical knowledge}), which models the domain-specific prior knowledge structure and is pre-constructed from the training corpus\footnote{For conciseness, in this paper, the \textit{prior working experience} and the \textit{prior medical knowledge} denote the retrieved radiology reports and the constructed medical knowledge graph, respectively.}. Finally, the MKD focuses on distilling the useful knowledge to generate proper reports. As a result, as shown in Figure~\ref{fig:introduction}, our PPKED has higher rate of accurately describing the rare and diverse abnormalities.

In summary, our main contributions are as follows:
\begin{itemize}
    \item In this paper, to alleviate the data bias problem, we propose the Posterior-and-Prior Knowledge Exploring-and-Distilling approach, which includes the Posterior and Prior Knowledge Explorer (PoKE and PrKE), and Multi-domain Knowledge Distiller (MKD).
    
    \item The PoKE explores posterior knowledge by employing the disease topic tags to capture the rare, diverse and important abnormal regions; The PrKE explores prior knowledge from prior working experience and prior medical knowledge; The MKD distills the extracted knowledge to generate reports.
    
    \item The experiments and analyses on the public IU-Xray and MIMIC-CXR datasets verify the effectiveness of our approach, which is able to outperform previous state-of-the-art model \cite{Chen2020Generating} on these two datasets.
    
\end{itemize}

The rest of the paper is organized as follows. Section \ref{sec:related_work} and Section \ref{sec:approach} introduce the related works and the proposed approach, respectively, followed by the experimental results (see Section \ref{sec:experiments}) and our conclusion (see Section \ref{sec:conclusion}).

\section{Related Works}
\label{sec:related_work}
The related works are introduced from three aspects: 1) Image Captioning, 2) Image Paragraph Generation and 3) Radiology Report Generation.

\subsection{Image Captioning}
The task of image captioning \cite{chen2015microsoft,Vinyals2015Show} has received extensive research interests. These approaches mainly adopt the encoder-decoder framework which translates the image to a \textit{single} descriptive sentence. Such framework have achieved great success in advancing the state-of-the-arts \cite{anderson2018bottom,liu2018simNet,lu2017knowing,rennie2017self,Xu2015Show,liu2019GLIED}. However, rather than only generating one single sentence, radiology report generation aims to generate a long paragraph, which consists of multiple structural sentences with each one focusing on a specific medical observation for a specific region in the radiology image.

\subsection{Image Paragraph Generation}
Beyond the traditional image captioning task, image paragraph generation that produces a long and semantic-coherent paragraph to describe the input image has recently attracted increasing research interests \cite{Krause2017Hierarchical,Liang2017Hierarchical,Yu2016Hierarchical}. To this end, a hierarchical recurrent network (HRNN) \cite{Krause2017Hierarchical,Liang2017Hierarchical} is proposed. In particular, the HRNN uses a two-level RNN model to generate the paragraph based on the image features extracted by a CNN. The two-level RNN includes a paragraph RNN and a sentence RNN, where the paragraph RNN is used to generate topic vectors and each topic vector is used by the sentence RNN to produce a sentence to describe the image. However, the correctness of generating abnormalities should be emphasized more than other normalities in a radiology report, while in a natural image paragraph each sentence has equal importance.

\subsection{Radiology Report Generation}

Writing a radiology report can be time-consuming and tedious for experienced radiologists, and error-prone for un-experienced radiologists \cite{Jing2018Automatic}. Similar to image paragraph generation, most existing works \cite{Jing2018Automatic,Xue2018Multimodal,Yuan2019Enrichment} attempt to adopt a HRNN to automatically generate a fluent report. However, due to the serious data deviation, these models are poor at finding visual groundings and are biased towards generating plausible but general reports without prominent abnormal narratives \cite{Jing2019Show,Li2018Hybrid,Yuan2019Enrichment,liu2021Contrastive}.

Currently, some approaches \cite{Jing2019Show,Li2018Hybrid,Yuan2019Enrichment,Zhang2020When,Li2020Auxiliary,liu2021Contrastive} have been proposed to alleviate data deviation. In detail, instead of only adopting a single sentence-level RNN to generate both the normal and abnormal sentences, \cite{Jing2019Show} introduced two RNNs as two different report writers, i.e., Normality Writer and Abnormality Writer, to help the model to generate more accurate normal and abnormal sentences, respectively. At the same time, \cite{Li2018Hybrid} proposed a hybrid model with template retrieval and text generation module, which focus on the generation of normal and abnormal sentences, respectively, to enhance the ability of model in describing abnormalities. Most recently, \cite{Zhang2020When} designed the medical graph based on prior knowledge from chest findings, in which each node is denoted by disease keywords representing one of the disease findings, so as to increase the capability of models to understand medical domain knowledge. Concurrently to our own work, the auxiliary signals introduced in \cite{Li2020Auxiliary} is similar to the idea of our approach. In particular, \cite{Li2020Auxiliary} only consider the medical graph, while we further leverage the disease topic tags and working experience to enhance the learning of posterior and prior knowledge, respectively. Besides, we further provides the evidence of this advantage of our approach on two public datasets.

It is observed that the data-driven RNNs designed in \cite{Jing2019Show} and \cite{Li2018Hybrid} could be easily misled by the rare and diverse abnormalities, disabling them from efficiently modeling the abnormal sentences.  
Different from them, our idea is mainly inspired by radiologists’ working patterns \cite{Li2019Knowledge}, to explore and distill the posterior and prior knowledge for accurate radiology report generation, which is missing in their approaches.
For the network structure, we first explore the posterior knowledge of input radiology image by proposing to explicitly extract the abnormal regions; Next, inspired by \cite{Li2018Hybrid} and \cite{Zhang2020When} which proved the effectiveness of retrieval module and medical knowledge graph, we leverage the retrieved reports and medical knowledge graph to model the prior working experience and prior medical knowledge. In particular, instead of only retrieving some sentences in previous works \cite{Li2018Hybrid,Li2019Knowledge}, we propose to retrieve a large amount of similar reports. Besides, since the templates may change over time, which was ignored in \cite{Li2018Hybrid}, using fixed templates will introduce inevitable errors. As a result, we treat the retrieved reports as latent guidance. In all, we combine the merits of retrieval module and knowledge graph in a single model. Finally, by distilling the useful prior and posterior knowledge, our approach could generate accurate reports.

\begin{figure*}[t]

\centering
\begin{center}
\includegraphics[width=0.85\linewidth]{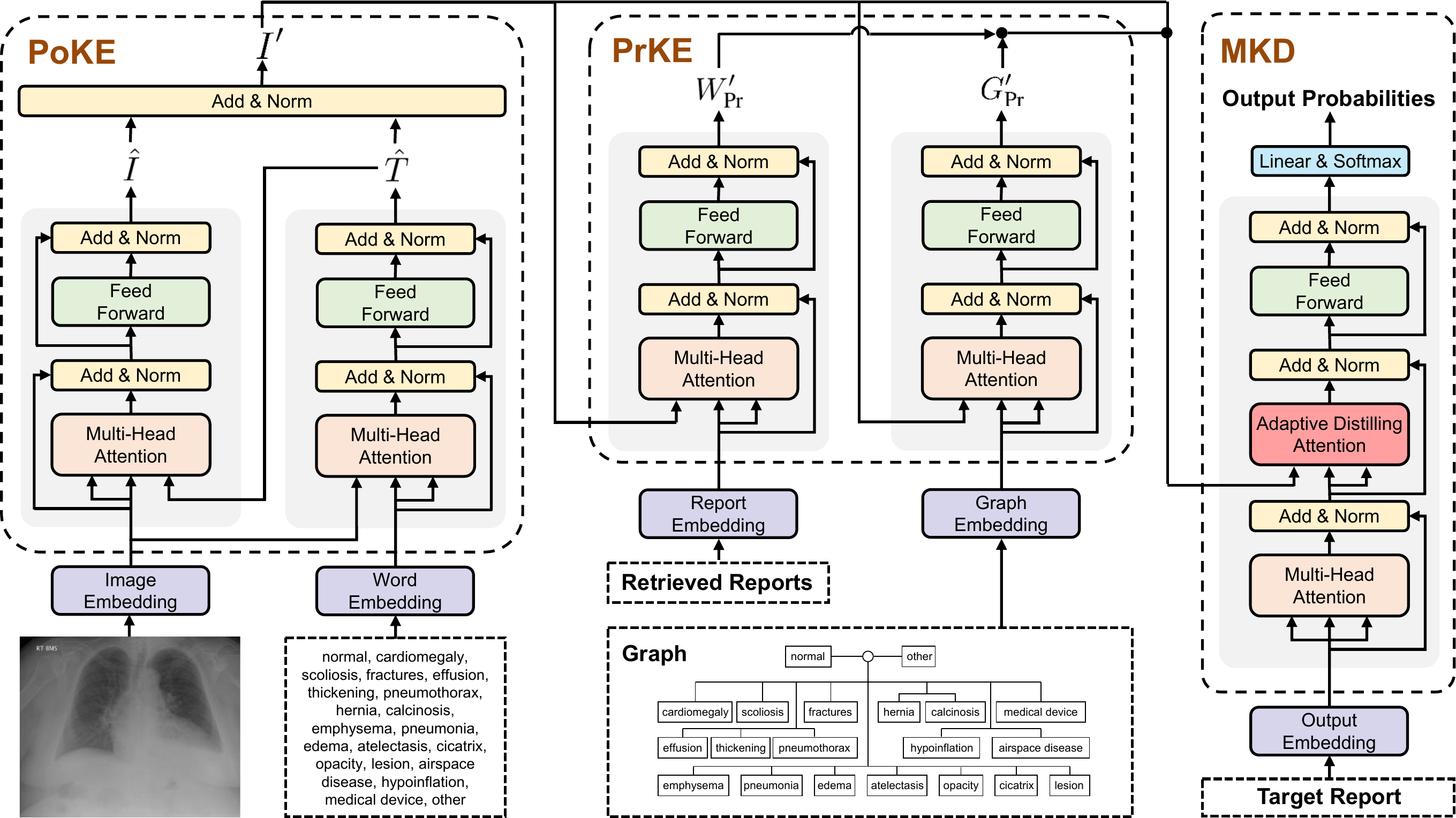}
\end{center}
\caption{Illustration of our proposed Posterior-and-Prior Knowledge Exploring-and-Distilling (PPKED) approach, which includes Posterior Knowledge Explorer (PoKE), Prior Knowledge Explorer (PrKE) and Multi-domain Knowledge Distiller (MKD). Specifically, PoKE explores the posterior knowledge by extracting the explicit abnormal regions and PrKE explores the relevant prior knowledge for the input image. At last, MKD distills accurate posterior and prior knowledge and adaptively merging them to generate accurate reports.}
\label{fig:approach}

\end{figure*}

\section{Posterior-and-Prior Knowledge Exploring-and-Distilling (PPKED)}
\label{sec:approach}

We first describe the background of PPKED and then introduce its three core components.

\subsection{Backgrounds}
The backgrounds are introduced from 1) Problem Formulation; 2) Information Sources and 3) Basic Module.

\smallskip\noindent\textbf{Problem Formulation} \
Given a radiology image encoded as $I$, we aim to generate a descriptive radiology report $R=\{y_1,y_2,\dots,y_{N_\text{R}}\}$. As shown in Figure~\ref{fig:approach}, our PPKED introduces a Posterior Knowledge Explorer (PoKE), a Prior Knowledge Explorer (PrKE) and a Multi-domain Knowledge Distiller (MKD). Specifically, we introduce the fixed topic bag $T$ that covers the $N_\text{T}$ most common abnormalities or findings to help the PoKE to explore the abnormal regions. The reason is that when radiologists examine the abnormal regions, they usually assign the disease topic tags to the abnormal regions. We also introduce the Prior Working Experience $W_\text{Pr}$ and the Prior Medical Knowledge $G_\text{Pr}$ extracted from the training corpus into our PrKE. Finally, the MKD devotes on distilling the useful information to generate reports, which can be formulated as:
\begin{equation}
\begin{aligned}
\small
&\text{PoKE}  :  \{I, T\} \to I' ; \\
&\text{PrKE} :  \{I', W_\text{Pr}\} \to W'_\text{Pr};
\quad \{I', G_\text{Pr}\} \to G'_\text{Pr} \\
&\text{MKD} : \{I', W'_\text{Pr}, G'_\text{Pr}\} \to R .
\end{aligned}
\end{equation}
In brief, the proposed PPKED takes $I, T, W_\text{Pr}, G_\text{Pr}$ as input to generate the robust report $R$.

\smallskip\noindent\textbf{Information Sources} \
We now describe how to obtain and encode the $I, T, W_\text{Pr}, G_\text{Pr}$ from training corpus in detail\footnote{Note that all encoded features have been projected by a linear transformation layer into the dimension of $d = 512$ in this paper.}.

$I$: Following \cite{Xue2018Multimodal,Yuan2019Enrichment,Huang2019Multi_Attention,Jing2019Show,Wang2018TieNet,liu2021Contrastive}, we adopt the ResNet-152 \cite{he2016deep} to extract 2,048 $7 \times 7$ image feature maps which are further projected into 512 $7 \times 7$ feature maps, resulting $I = \{i_1,i_2,\dots,i_{N_\text{I}}\} \in \R^{N_\text{I} \times d}$ ($N_\text{I} = 49$, $d = 512$).

$T$: In implementation, we choose $N_\text{T} = 20$ most common (abnormality) topics or findings, i.e., \textit{cardiomegaly}, \textit{scoliosis}, \textit{fractures}, \textit{effusion}, \textit{thickening}, \textit{pneumothorax}, \textit{hernia}, \textit{calcinosis}, \textit{emphysema}, \textit{pneumonia}, \textit{edema}, \textit{atelectasis}, \textit{cicatrix}, \textit{opacity}, \textit{lesion}, \textit{airspace disease}, \textit{hypoinflation}, \textit{medical device},  \textit{normal}, and  \textit{other}. We represent the topic bag with a set of vectors: $T=\{t_1,t_2,\dots,t_{N_\text{T}}\} \in \R^{N_\text{T} \times d}$, where $t_i \in \R^{d}$ refers to the word embedding of the $i^\text{th}$ topic.

$W_\text{Pr}$: To obtain the Prior Working Experience, we first extract the image embeddings of all training images from the last average pooling layer of ResNet-152. Then, given an input image, we again use the ResNet-152 to obtain the image embedding. At last, we retrieve $N_\text{K}=100$ images from the training corpus with the highest cosine similarity to the input image. The reports of the top-$N_\text{K}$ retrieved images are returned and encoded as the $W_\text{Pr} = \{R_1,R_2,\dots,R_{N_\text{K}}\} \in \R^{N_\text{K} \times d}$. In implementations, we use a BERT encoder \cite{Devlin2019BERT,Reimers2019Sentence-BERT} followed by a max-pooling layer over all output vectors as the report embedding module to get the embedding $R_i \in \R^{d}$ of the $i^\text{th}$ retrieved report.

$G_\text{Pr}$: In implementations, we follow \cite{Zhang2020When} to construct and initialize the medical knowledge graph. Specifically, based on the training corpus, for all images, we first build a universal graph $G_\text{Uni}  = (V, E)$, which models the domain-specific prior knowledge structure. In detail, we compose a graph that covers the most common abnormalities or findings. In particular, we use the common topics in the topic bag $T$. These $N_\text{T}$ common topics in $T$ are defined as nodes $V$ and are grouped by the organ or body part that they relate to. For topics grouped together, we connect their nodes with bidirectional edges, resulting in closely connected related topics. After that, guided by the input image $I$, we can acquire a set of nodes $V' = \{v'_1,v'_2,\dots,v'_{N_\text{T}}\} \in \R^{N_\text{T} \times d}$ encoded by a graph embedding module, which is based on the graph convolution operation \cite{Kipf2017GCN}. We regard the encoded $V'$ as the  prior knowledge $G_\text{Pr} \in \R^{N_\text{T} \times d}$.
Due to space limit, please refer to \cite{Zhang2020When} for the detailed description of medical knowledge graph.

\smallskip\noindent\textbf{Basic Module} \
We implement the proposed method upon the Multi-Head Attention (MHA) and Feed-Forward Network (FFN) \cite{Vaswani2017Transformer}. The MHA consists of $n$ parallel heads and each head is defined as a scaled dot-product attention:
\begin{align}
&\text{Att}_i(X,Y) = \text{softmax}\left(\frac{X\text{W}_i^\text{Q}(Y\text{W}_i^\text{K})^T}{\sqrt{{d}_{n}}}\right)Y\text{W}_i^\text{V} \nonumber \\
& \text{MHA}(X,Y) = [\text{Att}_1(X,Y); \dots; \text{Att}_n(X,Y)]\text{W}^\text{O}
\end{align}
where $X \in \R^{l_x \times d}$ and $Y \in \R^{l_y \times d}$ denote the Query matrix and the Key/Value matrix, respectively; $\text{W}_i^\text{Q}, \text{W}_i^\text{K}, \text{W}_i^\text{V} \in \R^{d \times d_n}$ and $\text{W}^\text{O} \in \R^{d \times d}$ are learnable parameters, where ${d}_{n} = d / {n}$. $[\cdot;\cdot]$ stands for concatenation operation.

Following the MHA is the FFN, defined as follows:
\begin{align}
\text{FFN}(x) = \max(0,x\text{W}_\text{f}+\text{b}_\text{f})\text{W}_\text{ff}+\text{b}_\text{ff} 
\end{align}
where $\max(0,*)$ represents the ReLU activation function; $\text{W}_\text{f} \in \R^{d \times 4d}$ and $\text{W}_\text{ff} \in \R^{4d \times d}$ denote  learnable matrices for linear transformation; $\text{b}_\text{f}$ and $\text{b}_\text{ff}$ represent the bias terms. It is worth noting that both the MHA and FFN are followed by an operation sequence of dropout \cite{srivastava2014dropout}, residual connection \cite{he2016deep}, and layer normalization \cite{ba2016layernormalization}.

\smallskip\noindent\textit{\textbf{Motivation}:} \ The MHA computes the association weights between different features.
The attention mechanism allows probabilistic many-to-many relations instead of monotonic relations, as in \cite{liu2019Aligning,Xu2015Show,Vaswani2017Transformer}. Therefore, we apply MHA to correlate the posterior and prior knowledge for the input radiology image, as well as distilling useful knowledge to generate accurate reports.

\subsection{Posterior Knowledge Explorer (PoKE)}
The PoKE is responsible for extracting the posterior knowledge from the input image, i.e., abnormal regions. To this end, the PoKE is conducted as (see Figure~\ref{fig:approach}):
\begin{align}
\label{eq:PoKE}
\hat{T} = \text{FFN}(\text{MHA}(I, T)) ; \ \hat{I} = \text{FFN}(\text{MHA}(\hat{T}, I))
\end{align}
In Eq.~(\ref{eq:PoKE}), the image features $I \in \R^{N_\text{I} \times d}$ are first used to find the most relevant topics and filter out the irrelevant topics, resulting in $\hat{T} \in \R^{N_\text{I} \times d}$. Then the attended topics $\hat{T}$ are further used to mine topic related image features $\hat{I} \in \R^{N_\text{I} \times d}$. Since $T$ contains the abnormality topics, the topic related image features can be referred as the abnormal regions. In this way, we can not only obtain the abnormal regions, but also align the attended abnormal regions with the relevant topics \cite{liu2019Aligning}, which imitates the working patterns of radiologists to assign the disease topic tags to the abnormal regions when examining the abnormal regions. It is worth noting that if we change the order from $I\to \hat{T}\to \hat{I}$ to $T\to \hat{I}\to \hat{T}$, the performance will drop. The reason is presumably due to the noisy topics, which contains a large amount of irrelevant topics in $T$, thus we should first filter out the irrelevant topics as the presented in Eq.~(\ref{eq:PoKE}).

Finally, since $\hat{I}$ and $\hat{T}$ are aligned, we directly add them up to acquire the posterior knowledge of the input image:
\begin{align}
   I' =  \text{LayerNorm}(\hat{I} + \hat{T})
\end{align}
where the LayerNorm denotes the Layer Normalization \cite{ba2016layernormalization}. Analogical to the process of how radiologists examine radiology images, we refer the acquired $I'$ to the first impression of radiologists after check the abnormal regions.

\subsection{Prior Knowledge Explorer (PrKE)}
The PrKE consists of a Prior Working Experience component and a Prior Medical Knowledge component. Both components obtain prior knowledge from existing radiology report corpus and represent them as $W_\text{Pr}$ and $G_\text{Pr}$ respectively. By processing $I'$ through these two components, we can acquire $W'_\text{Pr}$ and $G'_\text{Pr}$ which represent the prior knowledge relating to the abnormal regions of the input image. In implementation, we regard the abnormal features $I'$ as the lookup matrix. According to the attention theorem, the $I' \in \R^{N_\text{I} \times d}$ is the Query, and the $W_\text{Pr} \in \R^{N_\text{K} \times d}$/$G_\text{Pr} \in \R^{N_\text{T} \times d}$ is the Key and Value, which is defined as follows:
\begin{align}
W'_\text{Pr} &= \text{FFN}(\text{MHA}(I', W_\text{Pr})) \\
G'_\text{Pr} &= \text{FFN}(\text{MHA}(I', G_\text{Pr}))
\end{align}
Consequently, the results $W'_\text{Pr} \in \R^{N_\text{I} \times d}$ and $G'_\text{Pr} \in \R^{N_\text{I} \times d}$ turn out to be a set of attended (i.e., explored) prior knowledge related to the abnormalities of the input image, which could have potential to alleviate the textual data bias.

\subsection{Multi-domain Knowledge Distiller (MKD)}
After receiving the posterior and prior knowledge, the MKD performs as a decoder to generate the final radiology reports. For each decoding step $t$, the MKD takes the embedding of current input word $x_t = w_t + e_t$ as input ($w_t$: word embedding and $e_t$: fixed position embedding):
\begin{align}
h_t = \text{MHA}(x_t, x_{1:t})
\end{align}
Then, we employ the proposed Adaptive Distilling Attention (ADA) to distill the useful and correlated knowledge:
\begin{align}
h'_t = \text{ADA}(h_t, I', G'_\text{Pr}, W'_\text{Pr})
\end{align}
Finally, the $h'_t$ is passed to a FFN and a linear layer to predict the next word:
\begin{align}
y_{t} \sim p_{t}=\text{softmax}(\text{FFN}(h'_t)\text{W}_p + \text{b}_p)
\end{align}
where the $\text{W}_p$ and $\text{b}_p$ are the learnable parameters. Given the ground truth report $R^*=\{y^*_1,y^*_2,\dots,y^*_{N_\text{R}}\}$, we can train the PPKED by minimizing the cross-entropy loss:
\begin{align}
\label{eq:loss}
L_{\text{CE}}(\theta)=-\sum_{i=1}^{N_\text{R}} \log \left(p_{\theta}\left(y_{i}^{*} \mid y_{1: i-1}^{*}\right)\right)
\end{align}

\smallskip\noindent\textbf{Adaptive Distilling Attention (ADA)} \
Intuitively, radiology report generation task aims to produce a report based on the source radiology image $I'$, supported with the prior working experience $W'_\text{Pr}$ and the prior medical knowledge $G'_\text{Pr}$. Thus, the $W'_\text{Pr}$ and $G'_\text{Pr}$ play an auxiliary role during the report generation. To this end, we propose the ADA to make the model adaptively learn to distill correlate knowledge:
\begin{equation}
\footnotesize
\label{eq:ADA}
\text{ADA}(h_t, I', G'_\text{Pr}, W'_\text{Pr}) = \text{MHA}(h_t, I' + \lambda_1 \odot G'_\text{Pr} + \lambda_2 \odot W'_\text{Pr}) \nonumber 
\end{equation}
\vspace{-20pt}
\begin{align}
\lambda_1, \lambda_2 = \sigma\left(h_t\text{W}_h \oplus \left(I'\text{W}_{I} +  G'_\text{Pr}\text{W}_{G} + W'_\text{Pr}\text{W}_{W}\right)\right)
\end{align}
where $\text{W}_h, \text{W}_{I}, \text{W}_{G}, \text{W}_{W} \in \R^{d \times 2}$ are learnable parameters. $\odot$, $\sigma$ and $\oplus$ denote the element-wise multiplication, the sigmoid function and the matrix-vector addition, respectively. The $\lambda_1, \lambda_2 \in [0,1]$ weight the expected importance of $G'_\text{Pr}$ and $W'_\text{Pr}$ for each target word, respectively.

\begin{table*}[t]
\centering
\footnotesize
\begin{center}
\begin{tabular}{@{}l l c c c c c c c c@{}}
\toprule

Dataset & Methods & Year & BLEU-1 & BLEU-2 & BLEU-3 & BLEU-4 & METEOR & ROUGE-L & CIDEr \\
\midrule
\multirow{7}{*}[-3pt]{MIMIC-CXR \cite{Johnson2019MIMIC}} 
& CNN-RNN \cite{Vinyals2015Show} & 2015 &0.299 &0.184 &0.121 &0.084 &0.124 &0.263 & -  \\
& AdaAtt \cite{lu2017knowing} & 2017 &0.299 &0.185 &0.124 &0.088 &0.118 &0.266 & - \\
& Att2in \cite{rennie2017self} & 2017 &0.325 &0.203 &0.136 &0.096 &0.134 &0.276 & - \\ 
& Up-Down \cite{anderson2018bottom} & 2018 & 0.317 &0.195 &0.130 &0.092 &0.128 &0.267  & - \\
& Transformer \cite{Chen2020Generating}  & 2020 & 0.314 &0.192 &0.127 &0.090 &0.125 &0.265 & -\\
& R2Gen \cite{Chen2020Generating} & 2020 &0.353 &0.218 &0.145 &0.103 &0.142 &0.277 & - \\ \cmidrule(l){2-10}
& PPKED & Ours & \bf 0.360 & \bf 0.224 & \bf 0.149 & \bf 0.106 & \bf 0.149 & \bf 0.284 & \bf 0.237  \\
\midrule [\heavyrulewidth]

\multirow{8}{*}[-3pt]{IU-Xray \cite{Dina2016IU-Xray}} 
& HRNN \cite{Krause2017Hierarchical} & 2017 &0.439 & 0.281 & 0.190 & 0.133 & - & 0.342 & 0.261 \\
& CoAtt \cite{Jing2018Automatic} & 2018 & 0.455 & 0.288 & 0.205 & 0.154 & - & 0.369 & 0.277\\
& HRGR-Agent \cite{Li2018Hybrid} & 2018 &0.438 & 0.298 & 0.208 & 0.151 & - & 0.322 & 0.343\\
& CMAS-RL \cite{Jing2019Show} & 2019 &0.464 &0.301 &0.210 &0.154 & - &0.362  &0.275 \\
& SentSAT+KG \cite{Zhang2020When} & 2020 &0.441 &0.291 &0.203 &0.147 & - &0.367 & 0.304 \\
& Transformer \cite{Chen2020Generating} & 2020 & 0.396 & 0.254 & 0.179 & 0.135 & 0.164  & 0.342 & - \\
& R2Gen \cite{Chen2020Generating} & 2020 &0.470 &0.304 &0.219 &0.165 & 0.187 & 0.371  & - \\
\cmidrule(l){2-10}
& PPKED & Ours
& \bf 0.483 & \bf 0.315 & \bf 0.224 & \bf 0.168 & \bf 0.190 & \bf 0.376 & \bf 0.351
  \\ 
\bottomrule
\end{tabular}
\end{center}
\caption{Performance of the proposed PPKED and other state-of-the-art methods on the MIMIC-CXR and IU-Xray datasets. Higher value denotes better performance in all columns.
\label{tab:main_result}}
\end{table*}

\section{Experiments}
\label{sec:experiments}
In this section, we firstly describe two public datasets as well as some widely-used metrics and experimental settings in detail. Then we present the evaluation and analysis of the proposed approach.

\subsection{Datasets, Metrics and Settings}

We conduct the experiments on two public datasets, i.e., IU-Xray  \cite{Dina2016IU-Xray} and MIMIC-CXR \cite{Johnson2019MIMIC}.

\smallskip\noindent\textbf{IU-Xray} \
The IU-Xray \cite{Dina2016IU-Xray} is a widely-used benchmark dataset to evaluate the performance of radiology report generation methods. It contains 7,470 chest Xray images associated with 3,955 radiology reports. 
For data preparation, we first exclude the entries without the findings section and are left with 6,471 images and 3,336 reports. Then, following \cite{Jing2019Show,Li2019Knowledge,Li2018Hybrid,liu2021Contrastive}, we randomly split the dataset into 70\%-10\%-20\% training-validation-testing splits. There is no overlap of patients across train, validation and test sets. At last, we preprocess the reports by tokenizing, converting to lower-cases and removing non-alpha tokens. The top 1,200 words, which cover over 99.0\% word occurrences in the dataset, are included in our vocabulary.

\smallskip\noindent\textbf{MIMIC-CXR} \
We further adopt a recently released largest dataset to date, i.e., MIMIC-CXR \cite{Johnson2019MIMIC}, to verify the effectiveness of our approach. The dataset includes 377,110 chest X-ray images and 227,835 reports. Following \cite{Chen2020Generating} for a fair comparison, we use the official splits to report our results. As a result, the MIMIC-CXR dataset is split into 368,960 images/222,758 reports for training, 2,991 images/1,808 reports for validation and 5,159 images/3,269 reports for testing. We convert all tokens of reports to lower-cases and remove the tokens whose frequency of occurrence is less than 10, resulting in around 4k words.

\smallskip\noindent\textbf{Metrics} \
We adopt the widely-used BLEU \cite{papineni2002bleu}, METEOR \cite{Banerjee2005METEOR}, ROUGE-L \cite{lin2004rouge} and CIDEr \cite{vedantam2015cider}, which are calculated by the standard evaluation toolkit \cite{chen2015microsoft}. In particular, BLEU \cite{papineni2002bleu} and METEOR \cite{Banerjee2005METEOR} are originally proposed for machine translation evaluation. ROUGE-L~\cite{lin2003automatic} is designed for measuring the quality of summaries. CIDEr \cite{vedantam2015cider} is designed to evaluate image captioning systems.

\smallskip\noindent\textbf{Settings} \
We extract image features from both datasets with a ResNet-152 \cite{he2016deep}, which is pretrained on ImageNet and fine-tuned on CheXpert dataset \cite{Irvin2019CheXpert}. The extracted features are 2,048 feature maps in the shape of $7 \times 7$ which are further projected into 512 feature maps, i.e. $N_\text{I}$ is 49 and $d$ is 512. According the performance on the validation set, the number of retrieved reports $N_\text{K}$ and heads in MHA $n$ are set to 100 and 8, respectively. During training, following \cite{Zhang2020When,Jing2018Automatic}, we first pre-train our approach with a multi-label classification network and employ a weighted binary cross entropy loss for tag classification. Then we apply the Eq.~(\ref{eq:loss}) to train our full model. We use the Adam optimizer~\cite{kingma2014adam} with a batch size of 16 and a learning rate of 1e-4 for parameter optimization. 

\subsection{Main Results}
We compare our approach with a wide range of state-of-the-art radiology report generation models, i.e., HRNN \cite{Krause2017Hierarchical}, CoAtt \cite{Jing2018Automatic}, HGRG-Agent \cite{Li2018Hybrid}, CMAS-RL \cite{Jing2019Show}, SentSAT+KG \cite{Zhang2020When}, Transformer \cite{Chen2020Generating} and R2Gen \cite{Chen2020Generating}, as well as four image captioning models, namely CNN-RNN \cite{Vinyals2015Show}, AdaAtt \cite{lu2017knowing}, Att2in \cite{rennie2017self} and Up-Down \cite{anderson2018bottom}. For the IU-Xray dataset, except that the results of HRNN are implemented by ourselves, for systems designed for radiology report generation, we report the results from the original papers; For the MIMIC-CXR dataset, we directly cite the results from \cite{Chen2020Generating}. As shown in Table~\ref{tab:main_result}, our PPKED outperforms state-of-the-art methods across all metrics on both MIMIC-CXR and IU-Xray datasets. The improved performance of PPKED demonstrate the validity of our practice in exploring and distilling posterior and prior knowledge for radiology report generation.

\begin{table*}[t]

\footnotesize

\begin{center}
\begin{tabular}{@{}c c c c c c c|c c c c c c@{}}
\toprule
\multirow{2}{*}[-3pt]{Sections} & \multirow{2}{*}[-3pt]{Settings}  &\multirow{2}{*}[-3pt]{PoKE} & \multicolumn{2}{c}{PrKE}  & \multicolumn{2}{c|}{MKD} & \multicolumn{6}{c}{Dataset: IU-Xray \cite{Dina2016IU-Xray}}     \\ 
\cmidrule(lr){4-5} \cmidrule(lr){6-7} \cmidrule(lr){8-13} & & & PrMK & PrWE & TD & ADA & BLEU-1 & BLEU-2 & BLEU-3 & BLEU-4 & ROUGE-L & CIDEr  \\ \midrule [\heavyrulewidth]

\multirow{2}{*}{\ref{sec:effect_PoKE}}  & Base  & & & & &
&0.439 & 0.281 & 0.190 & 0.133 & 0.342 & 0.261
\\

& (a)  & $\surd$ & & & & & 0.449 & 0.294 & 0.199 & 0.144 & 0.353 & 0.285   \\ \midrule

\multirow{3}{*}{\ref{sec:effect_PrKE}} & (b)  & & $\surd$ & & & 
& 0.441 & 0.284 & 0.195 & 0.136 & 0.345 & 0.288 \\

& (c)  & & & $\surd$ & & 
& 0.449 & 0.288 & 0.195 & 0.146 & 0.346 & 0.296  \\

& (d)  & & $\surd$ & $\surd$ & & 
& 0.446 & 0.287 & 0.197 & 0.149 & 0.349 & 0.304   \\ \midrule

\multirow{3}{*}{\ref{sec:effect_MKD}} & (e)  & $\surd$ & $\surd$ & $\surd$ & & 
& 0.458 & 0.293 & 0.203 & 0.150 & 0.355 & 0.311  \\

& (f)  & $\surd$ & $\surd$ & $\surd$ & $\surd$ & 
& 0.476 & 0.309 & 0.222 & 0.165 & 0.372 & 0.337   \\

& PPKED  & $\surd$ & $\surd$ & $\surd$ & $\surd$ & $\surd$ 
 & \bf 0.483 & \bf 0.315 & \bf 0.224 & \bf 0.168 & \bf 0.376 & \bf 0.351
\\
\bottomrule
\end{tabular}
\end{center}
\caption{Quantitative analysis of our method. The Base model is the implementations of HRNN \cite{Krause2017Hierarchical} models.}
\label{tab:ablation}
\end{table*}

\begin{table}[t]
    \centering
    \footnotesize  
    \setlength{\tabcolsep}{2pt}
    \begin{center}
    \begin{tabular}{@{}l c c c c c c@{}}
        \toprule
        Categories & \begin{tabular}[c]{@{}c@{}}  TieNet \cite{Wang2018TieNet} \end{tabular}  & \begin{tabular}[c]{@{}c@{}}  DenseNet \cite{Irvin2019CheXpert} \end{tabular}  & \begin{tabular}[c]{@{}c@{}}  DenseNet+KG \cite{Zhang2020When} \end{tabular} & PPKED \\ \midrule [\heavyrulewidth]
        Emphysema & 0.79 & 0.89 & 0.89 & \bf 0.91 \\
        Pneumonia & 0.73 & 0.84 & 0.86 & \bf 0.87 \\ 
        Cardiomegaly & 0.85 & 0.87 & 0.91 & \bf 0.92\\
        Pneumothorax & 0.71 & 0.82 & 0.84 &  \bf 0.85 \\
        Lesion & 0.66 & 0.60 & 0.64 & \bf 0.69 \\ \midrule
        Normal & 0.75 & 0.80 & 0.81 & \bf 0.83 \\
        Average & 0.78 & 0.78 & 0.79 & \bf 0.80 \\
        \bottomrule
    \end{tabular}
    \end{center}
    \caption{Evaluation of abnormality classification results (AUCs).
    \label{tab:classification}}
\end{table}

\subsection{Quantitative Analysis}
In this section, we conduct the quantitative analysis to investigate the contribution of each component in our PPKED. 

\vspace{-10pt}
\subsubsection{Effect of Posterior Knowledge Explorer} \label{sec:effect_PoKE}
\vspace{-5pt}
Comparing the results of Base and (a) in Table~\ref{tab:ablation}, we can find that the incorporating Posterior Knowledge Explorer (PoKE) substantially boosts the performance of base model, e.g., 0.261 $\rightarrow$ 0.285 in CIDEr score. More encouragingly, the ``Base w/ PoKE'' even achieves competitive results with the state-of-the-art models on IU-Xray dataset (see Table~\ref{tab:main_result}), which demonstrates the effectiveness of our PoKE. We hypothesize that this performance gain may due to that PoKE can provide more accurate abnormal visual regions, which alleviate the visual data deviation problem. To verify this hypothesis, following \cite{Zhang2020When,Wang2018TieNet}, we randomly select five abnormality categories, i.e., `Emphysema', `Pneumonia', `Cardiomegaly', `Pneumothorax' and `Lesion' from the IU-Xray dataset, to test the models' ability of detecting abnormalities. As we can see, Table~\ref{tab:classification} proves our argument and verifies that PoKE can better recognize abnormalities which is important in clinical diagnosis.

\vspace{-10pt}
\subsubsection{Effect of Prior Knowledge Explorer} \label{sec:effect_PrKE}
\vspace{-5pt}
In this section, we evaluate the proposed two components, i.e., Prior Medical Knowledge (PrMK) and Prior Working Experience (PrWE), of Prior Knowledge Explorer (PrKE).

Table~\ref{tab:ablation} (b,c) shows that both the PrMK and PrWE can boost the performance, which prove the effectiveness of our approach. In detail, the PrMK can help the model learn enriched medical knowledge of the most common abnormalities or findings. For the PrWE, it significantly outperforms the Base model, which verifies the effectiveness of introducing existing similar reports.

By comparing the results of (b) and (c), we can find that the PrWE brings more improvements than the PrMK. We speculate the reason is that there are many similar sentences used to describe the normal regions in ground truth reports. Therefore, the description of normal regions can benefit from PrWE, especially when the appearance of normal reports dominate the whole dataset. It also shows that learning conventional and general writing style of radiologists is as important as accurately detecting abnormalities in radiology report generation.

Overall, since the PrMK and PrWE can improve the performance from different perspectives, combining PrMK and PrWE can lead to an overall improvement (see setting (d)). At the same time, PoKE and PrKE are able to alleviate the visual and textual data biases, respectively. The advantages of PoKE and PrKE can be united (see setting (e)).

\begin{table}[t]
    \centering
    \footnotesize  
    \setlength{\tabcolsep}{4pt}
    \begin{center}
    \begin{tabular}{@{}l c c c c c@{}}
        \toprule
        \multirow{2}{*}[-3pt]{Normality}  & $\lambda_1$ ($G'_\text{Pr}$) & $\lambda_2$ ($W'_\text{Pr}$) & \multirow{2}{*}[-3pt]{Abnormality}  & $\lambda_1$ ($G'_\text{Pr}$) & $\lambda_2$ ($W'_\text{Pr}$) \\
        \cmidrule(lr){2-3} \cmidrule(l){5-6}  &  0.27 & 0.44 & & 0.81 & 0.63  \\
        \bottomrule
    \end{tabular}
    \end{center}
    \caption{The analysis of our proposed Adaptive Distilling Attention. We report the average distilling values $\lambda_1$ and $\lambda_2$ in Eq.~(\ref{eq:ADA}) according to the sentences which describe the \textit{normalities} and \textit{abnormalities} in the radiology images.
    \label{tab:ADA}}
\end{table}

\begin{figure*}[t]

\centering
\begin{center}
\includegraphics[width=1\linewidth]{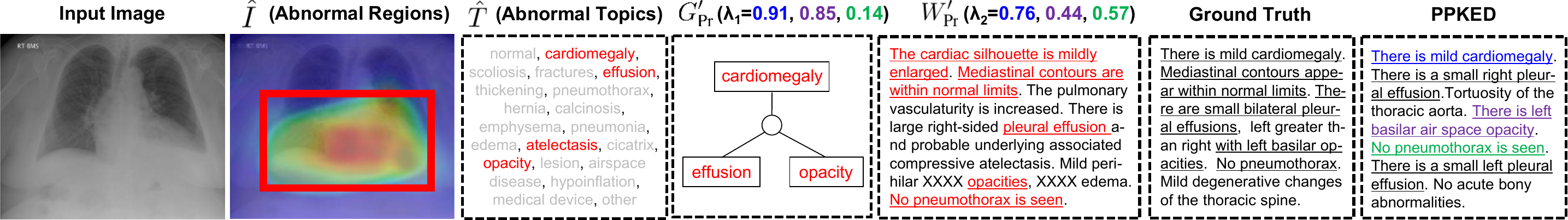}
\end{center}
\caption{We give the visualization of the PPKED. Please view in color. The Red bounding box and Red colored text denote the knowledge explored (i.e., attended) by our approach; For $W'_\text{Pr}$, we show the retrieved report with highest attention weight; For $G'_\text{Pr}$, we show the nodes whose attention weights exceeds 0.2. The Blue, Purple and Green colored numbers in brackets denote the distilling weight values in our Adaptive Distilling Attention of the Multi-domain Knowledge Distiller when generating corresponding sentences. Underlined text denotes alignment between the ground truth text and generated/retrieved text.}
\label{fig:example}
\end{figure*}

\vspace{-10pt}
\subsubsection{Effect of Multi-domain Knowledge Distiller} \label{sec:effect_MKD}
\vspace{-5pt}
In implementation, our MKD is based on the Transformer Decoder (TD) \cite{Vaswani2017Transformer} equipped with the proposed Adaptive Distilling Attention (ADA).

The lower part of Table~\ref{tab:ablation} illustrates that the model with a LSTM-based Decoder (e) performs worse than model with a Transformer-based Decoder (f), which directly demonstrates the effectiveness of the Transformer Decoder can effectively model long sequences.

Moreover, as shown in the last line of Table~\ref{tab:ablation}, it is clear that our Adaptive Distilling Attention (ADA) successfully boosts the performance, verifying the effectiveness of our approach. To further understand ADA’s ability of adaptively distilling useful prior and posterior knowledge, we summarize the average distilling weight values $\lambda_1$ and $\lambda_2$ according to the sentence type (normality and abnormality) in Table~\ref{tab:ADA}. Specifically, following \cite{Jing2018Automatic}, we consider sentences which contain ``no'', ``normal'', ``clear'', ``stable'' as sentences describing normalities. As expected, the values of $\lambda_1$ and $\lambda_2$ generating the abnormalities are larger than the values generating the normalities. The reason is that both $G'_\text{Pr}$ and $W'_\text{Pr}$ contains much knowledge about the abnormalities, which indicates our ADA are capable of learning to efficiently distill the explored prior and posterior knowledge.

It is also worth noting that since the retrieved reports in $W'_\text{Pr}$ contains the knowledge about the normalities, $\lambda_2$ is larger than $\lambda_1$ when generating the normalities. Therefore, in addition to distilling the knowledge about the abnormalities, our ADA can also capture the most related useful knowledge about the normalities for generating accurate normality sentences. The ability of distilling the accurate knowledge about the normalities can be verified by the best AUC score in terms of the `Normal' category in Table~\ref{tab:classification}, which proves our argument.

\subsection{Qualitative Analysis}
In Figure~\ref{fig:example}, we give an intuitive example to better understand our approach. As we can see, in PoKE, the original image features find the most relevant topics including the \textit{cardiomegaly}, \textit{effusion}, \textit{atelectasis} and \textit{opacity}, which then attend to the relevant abnormal regions, verifying the capability of PoKE to extract explicit abnormal visual regions (Red bounding box).

In particular, the PPKED generates structured and robust reports, which show significant alignment with ground truth reports and are supported by accurate abnormal descriptions as well as correspondence with the visualized abnormal regions. For example, the generated report correctly describes ``\textit{There is mild cardiomegaly}'', ``\textit{There is left basilar air space opacity}'' and ``\textit{There is a small right/left pleural effusion}''. In detail, 1) due to higher rate of involving explicit abnormal visual information provided by the PoKE, the generated report contains accurate abnormalities and locations and also share a well balance of normal sentences and abnormal sentences. This phenomenon shows that our approach can efficiently alleviate the visual data deviation problem. 2) The generated report and the explored prior knowledge show correspondence with the ground truth reports, e.g., \textit{cardiomegaly}, \textit{opacity} and \textit{effusion}, which verifies that PrKE is capable of exploring the accurate prior textual knowledge to efficiently alleviate the textual data bias; 3) The reasonable distilling weight values prove that the MKD is able to distill accurate information from the explored posterior and prior knowledge, and adaptively merging them for generating each accurate sentence.

In brief, the qualitative analysis proves our arguments and verify the effectiveness of our proposed approach in alleviating the data bias problem by exploring and distilling posterior and prior knowledge.

\section{Conclusion}
\label{sec:conclusion}
In this paper, we present an effective approach of exploring and distilling posterior and prior knowledge for radiology report generation. Our approach imitates the working patterns of radiologists to alleviate the data bias problem. The experiments and analyses on the MIMIC-CXR and IU-Xray datasets verify our arguments and demonstrate the effectiveness of our method. In particular, our approach not only generates meaningful and robust radiology reports supported with accurate abnormal descriptions and regions, but also outperforms previous state-of-the-art models on the two public datasets.

\section*{Acknowledgments}
This paper was partially supported by the IER foundation (No. HT-JD-CXY-201904) and Shenzhen Municipal Development and Reform Commission (Disciplinary Development Program for Data Science and Intelligent Computing). Special acknowledgements are given to Aoto-PKUSZ Joint Lab for its support. 
% We thank all the anonymous reviewers for their constructive comments and suggestions.
% Xian Wu and Yuexian Zou are the corresponding authors of this paper.

We sincerely thank all the anonymous reviewers and chairs for their constructive comments and suggestions that substantially improved this paper.

{\small
\bibliographystyle{ieee_fullname}
\bibliography{egbib}
}

\end{document}